\documentclass{Interspeech}

\interspeechcameraready 

\usepackage{todonotes}
\usepackage{subcaption}
\usepackage{arydshln}
\usepackage{pgfplots}
\usepackage{multirow}

\title{Loquacious Set: 25,000 Hours of Transcribed and Diverse English Speech Recognition Data for Research and Commercial Use}

\author[affiliation={}]{Titouan}{Parcollet}
\author[affiliation={}]{Yuan}{Tseng}
\author[affiliation={}]{Shucong}{Zhang}
\author[affiliation={}]{Rogier}{van Dalen}

\affiliation[nocounter]{AI Center Cambridge}{Samsung}{United Kingdom}

\email{\{t.parcollet,s1.zhang,r.vandalen\}@samsung.com}


\begin{document}

\maketitle

\begin{abstract} 
    Automatic speech recognition (ASR) research is driven by the availability of common datasets between industrial researchers and academics, encouraging comparisons and evaluations. LibriSpeech, despite its long success as an ASR benchmark, is now limited by its size and focus on clean, read speech, leading to near-zero word error rates. More recent datasets, including MOSEL, YODAS, Gigaspeech, OWSM, Libriheavy or People's Speech suffer from major limitations including licenses that researchers in the industry cannot use, unreliable transcriptions, incorrect audio data, or the lack of evaluation sets. This work presents the Loquacious Set, a 25,000-hour curated collection of commercially usable English speech. Featuring hundreds of thousands of speakers with diverse accents and a wide range of speech types (read, spontaneous, talks, clean, noisy), the Loquacious Set is designed to work for academics and researchers in the industry to build ASR systems in real-world scenarios.
\end{abstract}

\section{Introduction}

The success of deep learning in automatic speech recognition (ASR) hinges on access to standardised datasets. Corpus like the Wall Street Journal (WSJ) \cite{paul1992design} and TIMIT \cite{garofolo1993darpa} were crucial for early ASR research, but their limited size and price rapidly became limiting issues. The advent of LibriSpeech \cite{panayotov2015librispeech}, with its open CC-BY license and 1,000 hours of transcribed speech, revolutionized English ASR research. It provided a common benchmark for academics and industrial researchers, enabling the rapid advancement of ASR technologies. However, the widespread use of LibriSpeech has led to overfitting due to its simplicity. Indeed, LibriSpeech contains no background noise or spontaneous speech, with systems achieving near-zero word error rates while struggling to generalize to real-world scenarios. As a result, current state-of-the-art and ``open-source'' large-scale models such as Whisper \cite{radford2023robust}, Conformer-1 \cite{zhang2024conformer}, Canary and Parakeet have all been trained with privately held datasets making scientific comparisons with newly developed ASR methods almost impossible. Hence, the speech community needs a successor to LibriSpeech with the following characteristics: 1.\ Sufficiently large for training modern models, yet with defined smaller subsets for quick and cheaper experimentation; 2.\ A permissive license to support academic and industrial research; 3.\ Acoustic and linguistic diversity (noises, accents, read/spontaneous speech); 4.\ Normalised transcriptions with few errors for training; and 5.\ Simple data preparation and download with documentation. 

While numerous datasets have been proposed, none have met all the above criteria. Both MOSEL \cite{gaido2024mosel} and YODAS \cite{li2023yodas} are too large to be easily manipulated, do not offer well-defined smaller subsets and contain unreliable labels (or pseudo-labels) and acoustic data. For instance, we discovered that up to half of the English data from YODAS is not English (see section \ref{sec:dataset}). Libriheavy \cite{kang2024libriheavy} is like LibriSpeech but larger, containing only speech from audiobooks. People's Speech \cite{galvez2021people} is mostly limited to American English and suffers from significant noise in its transcriptions, rendering approximately 85\% of the data unreliable for training and evaluation. SpeechStew \cite{chan2021speechstew}, OWSM \cite{peng2023reproducing}, and Gigaspeech \cite{chen2021gigaspeech} are covered by non-commercial licenses or copyrights forbidding industrial research. Common Voice \cite{ardila2019common} is too small on its own. Finally, most of these datasets also do not offer any validation or test sets for proper benchmarking.

This article introduces the Loquacious Set, a comprehensive collection of 25,000 hours of annotated English speech. It is large enough to train large-scale modern models but also contains three smaller subsets for institutions with lower storage capacity. The licenses of the data allow commercial use and research. The data contain hundreds of thousands of speakers with a wide range of accents and speech types (e.g. read, spontaneous, talks, clean, noisy). This blend of clean and challenging audio samples is what makes the Loquacious Set a good solution for training and evaluating all-purpose ASR systems in real-world conditions as it does not specialise in a single use case but addresses many. In addition to data preparation scripts, we provide SpeechBrain \cite{speechbrain} ASR training recipes demonstrating how easily this dataset can be used.%
\footnote{\url{https://huggingface.co/datasets/speechbrain/LoquaciousSet}}

\begin{table*}[h!]
    \centering
    \caption{Data partitioning of the 6 subsets of the Loquacious set. ``Available hours'' reports the total audio duration available in the source dataset. The ``Small'', ``Medium'', ``Large'', ``Clean'', ``Dev.'' and ``Test'' columns describe how much is taken from each source dataset to compose the corresponding subset of the Loquacious set. The ``Clean'' subset does not have People's Speech and YODAS data as, even after verification, the transcriptions may remain unreliable.}
    \label{table:dataset}
    \begin{tabular}{l@{~~}l@{~~}c@{~~~}c@{~~~}|c@{~~~}c@{~~~}c@{~~~}c@{~~~}c@{~~~}c@{~~~}c@{~~}}
    \toprule
     \textbf{Source dataset} & \textbf{Speech type} & \textbf{Available hours} & \textbf{License} &\textbf{small} & \textbf{medium} & \textbf{large} & \textbf{clean} &  \textbf{dev} & \textbf{test}\\
     \midrule
     VoxPopuli& Talks, non-native & 550 & CC0  & 50 & 500 & 550 & 550 &  5  & 5   \\
     Common Voice 18.0 & Read, clean, noisy & 1,600 & CC0  & 50  & 500  & 1,600 & 1,600 & 5 & 5  \\
      & native, non-native, far-field &  &   &  &   &  &  &  &   \\
     People's Speech & Read, talks, spontaneous & 30,000 &  CC BY 4.0  & 50  & 500 & 5,900 & 0 & 0 &  0 \\
     & native, non-native &  &   &  &   &  &  &  &   \\
     & clean, noisy, far-field &  &   &  &   &  &  &  &   \\
     YODAS & YouTube mixed & 45,000 & CC BY 3.0 & 50 & 500 & 6,100 & 0 & 1.5 & 1.5  \\
     Libriheavy & Read, clean, native & 50,000 & CC BY 4.0  & 50 & 500 & 11,000 & 11,000 & 0 & 0 \\
     LibriSpeech & Read, clean, native & 960 & CC BY 4.0 & 0 & 0 & 0 & 0 & 5 & 5 \\
     \midrule
     \textbf{Total hours} & & \textbf{128,110} & & \textbf{250} & \textbf{2,500} & \textbf{25,150} & \textbf{13,150} & \textbf{16.5} & \textbf{16.5}  \\
    \bottomrule
    \end{tabular}
\end{table*}

\section{Large Scale English ASR Datasets}
\label{sec:related}

The Loquacious Set is not the first attempt to create a standardised large-scale ASR dataset for the English language. This section discusses others relevant initiatives.

The CommonVoice dataset \cite{ardila2019common} is based on crowd-sourcing, and can theoretically leverage millions of users recording each a few samples on their devices. While it only provides 1,600 hours of speech, CommonVoice shines for the diversity in accents and acoustic conditions originating from almost a hundred thousand different speakers. The licence allows commercial use. Unfortunately, CommonVoice suffers from its lack of standardisation, as new versions are released frequently, and training, validation and test sets are therefore changing continuously. Finally, the transcriptions are not normalised to ensure a fair comparison between ASR models.

GigaSpeech \cite{chen2021gigaspeech}, on the other hand, is a truly large-scale English ASR corpus by today's standard as it provides up to 10,000 hours of labelled speech. By crawling three different sources including audiobooks, podcasts and YouTube videos, GigaSpeech is able to provide a diverse set of linguistic and acoustic conditions with standardised text normalisation and carefully crafted validation and testing sets. Unfortunately, Gigaspeech suffers from data licenses preventing industrial research. Indeed, and while the data preparation script is licensed under Apache 2.0, the authors of the dataset do not own the copyright of the underlying data, and the licence of the YouTube videos, for instance, has not been checked for commercial use. SpeechStew \cite{chan2021speechstew} and OWSM \cite{peng2023reproducing} suffer from the same issue. 

Libriheavy \cite{kang2024libriheavy} is Librispeech but scaled up. It extends the dataset to 50,000 hours of labelled speech. The content mostly comes from the LibriVox project, a large collection of audiobooks with very clean speech. ASR systems trained on Libriheavy obtain state-of-the-art performance on the standard evaluation sets of Librispeech, but they fall short on spontaneous or noisy speech. Libriheavy is free to use and the data is covered by the permissive CC BY 4.0 licence. Multilingual Librispeech \cite{pratap2020mls} (MLS) is very similar to Libriheavy as it comes from the same source but is multilingual.

People's Speech \cite{galvez2021people} and YODAS \cite{li2023yodas} are the most recent attempts at overcoming the read versus spontaneous speech issue at large scale. The People's Speech provides 30,000 hours of mostly American English labelled speech originating from \textit{archive.org} (i.e. from the internet) excluding videos from YouTube. The vast majority of the data comes from American government sources or interviews. However, the People's Speech transcriptions suffer from misalignments between the audio and the text or even wrong transcriptions e.g. repetitions and filler words are sometimes transcribed, but sometimes not. YODAS exhibits the same issues but at an even larger scale. YODAS provides around 45,000 hours of manually transcribed English YouTube videos. Unfortunately, annotations are not normalised or checked, and the audio is not further validated. Upon investigation, we found that roughly 50\% of the English data was not English. Manual annotations on YouTube contain several irregularities including, but not limited to, emojis, textual descriptions of a scene or messages to the subscribers. MOSEL \cite{gaido2024mosel} is very similar to YODAS as it mostly contains speech from YouTube videos and suffers from the same issues.

All existing large-scale English ASR datasets have major limitations. The Loquacious set solves these issues by combining the whole or subsets of CommonVoice, VoxPopuli \cite{wang2021voxpopuli}, Libriheavy, People's Speech and YODAS into a single, curated and normalised large-scale corpus containing 25,000 hours of labelled English speech with allowed commercial use. 
\section{The Loquacious Set}
\label{sec:dataset}

This section describes the data selection and corpus structure of the Loquacious Set (section \ref{subsec:data_selection}) as well as the process used to normalise the transcriptions (section \ref{subsec:text_norm}).\\

\noindent\textbf{License considerations.} The Loquacious dataset uses existing data and does not impose any new licensing restrictions. Each component of the original datasets keeps its original Creative Commons license, all of which permit commercial use (see Table \ref{table:dataset}). The code to reproduce the dataset is distributed under the Apache 2.0 license. 

\subsection{Audio data selection and corpus structure}
\label{subsec:data_selection}

The Loquacious set contains four training splits and two evaluation sets: small (250 hours), medium (2,500 hours), large (25,000 hours), clean (13,000 hours), dev (16.5 hours) and test (16.5 hours). They originate from a combination of six datasets: CommonVoice, VoxPopuli, Libriheavy, People’s Speech, YODAS and LibriSpeech. The amount of data selected from each dataset and their license is given in Table \ref{table:dataset}.\\

\noindent\textbf{Training sets.} The small and medium sets are constructed by taking 50 and 500 hours respectively from each source dataset. Smaller sets are subsets of larger ones. Samples are picked uniformly provided that they fulfil the rules detailed thereafter. The clean subset is made of the datasets whose transcriptions have been manually checked before publication, i.e.\ CommonVoice, Voxpopuli and Libriheavy. As discussed later, the People's Speech and YODAS transcriptions, even after our text processing, may still contain errors. The large split contains data from all corpora up to a certain threshold to reach 25,000 hours of speech. \\

\noindent\textbf{Evaluation sets.} The Loquacious Set provides a development and a test set containing a bit more than 16.5 hours each. First, five hours are uniformly picked from the validation and test sets of CommonVoice, VoxPopuli and LibriSpeech ``dev-other'' or ``test-other'' to compose the clean base of our validation and test splits of our dataset. Spontaneous speech validation is done with YODAS. Hence, we extracted 3 hours of data from the \textit{en003} subset of YODAS. After manual verification and splitting, this yielded 1 and 1.5 hours for validation and testing, respectively. \\

\noindent\textbf{Audio processing.} All samples are resampled to 16khz and stored as bits in the dataset \textit{.parquet} files using the \textit{soundfile} library. Audio files are segmented to extract only the relevant piece of speech based on timestamps. As a result, each row of the dataset contains the metadata, the labels, as well as the binary data of the corresponding audio segment.\\

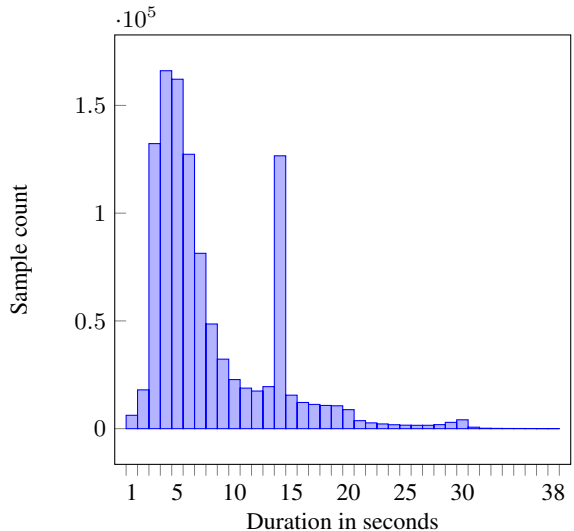
\begin{figure}[h!]
\begin{tikzpicture}
    \begin{axis}[
        ybar interval,
        xmin=0,
        xmax=40,
        x=0.15cm,
        xticklabels={1,,,,5,,,,,10,,,,,15,,,,,20,,,,,25,,,,,30,,,,,,,,38,,},
        xlabel=Duration in seconds,
        ylabel=Sample count,
        ytick pos=left,
        xtick pos=bottom,
        grid=none,
        fill=blue]
            \addplot table[x=bins, y=count, col sep=comma]{resources/frequency_large.csv};%
    \end{axis}
\end{tikzpicture}
\caption{ Durations of the 9.6M samples of the large training split of the Loquacious Set. The average is 9.6 seconds.}\label{fig:duration}
\end{figure}

\noindent\textbf{Audio filtering.} Samples longer than 40 seconds and shorter than one second are excluded from the dataset. For YODAS, we increased the lower bound to 3 seconds as most of the samples below this threshold correspond are just noises. The distribution of the sample duration on the large set is given in Figure \ref{fig:duration}. It contains 9.4 million samples with an average duration of 9.6 seconds. While most of the filtering happened at the text level, YODAS required one extra step. This dataset, created from YouTube videos, required language identification with SpeechBrain\footnote{\url{https://huggingface.co/speechbrain/lang-id-voxlingua107-ecapa}} due to a language mismatch between the transcriptions and the audio. We ran language identification on three English subsets (``en000'',  ``en001'' and ``en003''), identifying 9,200 hours of valid English speech out of 16,500 hours. A manual review of a few randomly sampled sentences confirmed these excluded samples were non-English or unintelligible speech.\\

\noindent\textbf{Acoustic conditions.} The Loquacious Set is a diverse dataset with acoustic conditions ranging from American accented read and clean speech to heavily accented spontaneous speech with babble noise and reverberation. Half of the dataset can be considered as clean speech as VoxPopuli and Libriheavy have clean recording conditions. VoxPopuli adds heavily accented speech to it (EU parliament with native and non-native speakers). The other half of the dataset can be considered as potentially noisy. Indeed, CommonVoice contains read speech from high-end microphones in calm rooms but also read speech from people in subways recording with their headphone microphones. YODAS and People's Speech originate from YoutuBe videos, meetings and interviews, and therefore offer spontaneous speech in a plethora of different acoustic conditions.

\begin{table*}[h!]
    \centering
    \caption{Speech recognition results of various sized conformer encoder-decoder trained on the different sets of the Loquacious Set. Word error rate percentages (i.e. lower is better) are obtained with joint CTC-Attention beam search decoding and no language model. The 480M model ($\dagger$) is trained with data augmentation.}
    \label{tab:res}
    \begin{tabular}{l@{~~}c@{~~}c@{~~}c@{~~}c@{~~}c@{~~}}
    \toprule
     \textbf{Training split.} & \textbf{Loquacious Set} & \textbf{Loquacious Set} & \textbf{Librispeech} & \textbf{CommonVoice} & \textbf{Voxpopuli}  \\
      \# Parameters & \textit{Dev.} & \textit{Test} & \textit{Test-other} & \textit{Test} & \textit{Test}  \\
     \midrule
     \textbf{Small (250 hours)} &  &  &  &  &   \\
     25M & 22.8 & 24.3 & 22.8  & 38.4 & 16.2  \\
     100M & 22.3  & 23.8 & 22.5 & 37.6 & 16  \\

     \textbf{Medium (2,500 hours)} &  &  &  &  &   \\
     25M & 13.3 & 14.3 & 12.3 & 23.9 & 9.7  \\
     100M & 11.3 & 12.4 & 10.1 & 20.5  & 8.8 \\
     250M & 10.7 & 11.9 & 9.8 & 20.0 &  7.9 \\

     \textbf{Large (25,000 hours)} &  &  &  &  &   \\
     100M & 8.4 & 9.3 & 6.0 & 15.9 & 7.9 \\
     250M & 7.9 & 8.8 & 5.7 & 14.6 & 7.6 \\
     480M$^\dagger$ & 6.8 & 7.5 & 4.6 & 12.0 & 6.9 \\

     \textbf{Clean (13,000 hours)} &  &  &  &  &   \\
     250M & 9.0 & 10.3 & 6.0 & 16.6 & 8.2 \\
     
    \bottomrule
    \end{tabular}
\end{table*}

\subsection{Transcriptions normalisation and metadata}
\label{subsec:text_norm}

Without any text normalisation, the correctness of the transcriptions varies a lot. For instance, Libriheavy follows the Librispeech normalisation and can be considered as very clean. On the contrary, YODAS has no normalisation at all, and transcriptions may contain symbols, numerical values, emojis, event descriptors like \textit{[applause]}, or anything that the author of the YouTube video may want to add. Hence, it is critical to unify the normalisation of the transcriptions. The Loquacious Set text normalisation process follows five steps.

First, numerals are converted to their equivalent string with the Nemo text-processing tool
\footnote{\url{https://github.com/NVIDIA/NeMo-text-processing}}
based on WFSTs. Then, we remove any sentence containing accented and non-English characters like \textit{ă} or \textit{œ} as pronunciation will be too uncertain (i.e. for Common Voice) or it may indicate non-English sentences. The next step is to remove all word boundaries and punctuation and wrongly encoded letters. The fourth step is to replace common symbols with their corresponding string such as \textit{dollar} for \textit{\$} or \textit{and} for \textit{\&}. Sentences with more than four symbols are removed as the pronunciation may again be too uncertain. The fifth step relies on the \textit{nltk} library to remove any remaining YouTube descriptors like \textit{[explosion]} or \textit{[light music]}. Finally, all words are upper-cased and checked to only contain the 26 Latin letters or the apostrophe. Due to imperfect audio alignment in YODAS and People's Speech, the speech at the beginning and end of an audio segment may only contain a portion of the corresponding word.\\

\noindent\textbf{Metadata.} It is not possible to obtain complete information about the sex or identity of the speakers due to the heterogeneity of the source datasets. Indeed, for some datasets like Libriheavy, each speaker is attributed a unique speaker identifier while this is not true for CommonVoice, YODAS or People's Speech. Hence, the Loquacious Set provides for each sample mandatory and optional information. Mandatory information include an audio identifier, the duration of the segment in seconds, the binary representation of the audio segment and the transcription. The optional metadata are the sex and the speaker identifier as given (i.e.\ unnormalised) in the original dataset. This heterogeneity prevents us from releasing any precise statistics on the number of speakers or their sex. However, as Common Voice already almost contains one hundred thousand speakers, it is fair to assume that the real number is well above this value.

\section{Speech Recognition Experiments}

This section details the experimental protocol (section \ref{subsec:exp_prot}) as well as the speech recognition results (section \ref{subsec:results}) obtained with the Loquacious Set on various benchmarking speech recognition datasets. 

\subsection{Experimental protocol}
\label{subsec:exp_prot}

Models are trained on the three subsets of the Loquacious Set, hence containing either 250, 2500 and 25,000 hours of training data. Evaluation is conducted with the validation and test sets of the Loquacious Set as well as the test-other partition from Librispeech and test sets from Voxpopuli and CommonVoice. \\

\noindent\textbf{ASR models and hyperparameters.} The standard conformer encoder-decoder architecture is used. Four different sizes of models containing 25, 100, 250 and 480 million parameters are proposed. The smallest one is made of 12 encoder blocks of dimension 256, while the three others contains 12, 14 and 18 blocks of dimensions 512, 768 and 1,024 respectively. The vocabulary size is set to 5,120 BPE tokens for the 2,500 and 25,000 hours partitions and 1,024 for the 250-hour one. Multi-task training with joint CTC/attention \cite{kim2017joint} is performed for 250,000 steps for the 250 and 2,500 hours subsets and 500,000 steps for the full 25,000 hours dataset. The Adam optimizer is coupled with a warmup and decay learning rate scheduler and 40,000 steps of warmup. Decoding is performed with joint CTC/Attention and no language model. Training is done on four Tesla V100 for the 250 hours and 2500 hours subsets while height of them are used for the full 25,000 hours dataset. The batch size per GPU is 150 seconds and a gradient accumulation factor of 3 is used in combination of the four GPUs and 1 when using eight. No data augmentation is used for most models except the 480M parameters with SpecAugment. The list of hyperparameters is given in the SpeechBrain recipe.

\subsection{Speech recognition results}
\label{subsec:results}

The obtained ASR performance are reported in Table \ref{tab:res}. First of all, and in line with the literature and overall consensus, it is fairly clear that both adding more data and increasing the size of the ASR model lead to reduced word error rates. For instance, on the test set of the Loquacious Set, the WER drops from 24.3\% to 7.5\% when switching from the small data split with 25M parameters conformer to the full dataset and 480M parameters. This trend is also observed in all available test sets. 

Secondly, it is interesting to note that not all test sets benefit equally from adding more data. In the case of VoxPopuli, the WER is almost halved (e.g. from 16\% to 8.8\% with 100M parameters) when going from 250 hours to 2,500 hours, while it only drops to 7.9\% when jumping to the full 25,000 hours of data. On the other hand, Librispeech sees much higher drops in WER, even though it still tends to reduce with the amount of data (e.g. 22.5\% to 10.1\% to 6.0\%). It may be because the 2500 hours already contain 500 hours of the 550 available hours of the VoxPopuli dataset, while this is to compare to 500 hours of Libriheavy against 11,000 hours in the large set. As expected, having proper in-domain data is the primary factor to improve WERs. However, adding closely related data also helps as WERs obtained on the test sets of VoxPopuli and CommonVoice are lower than what would be observed with an ASR system trained on each of them individually. For instance, the same 100M model trained and evaluated only with CommonVoice achieves a test WER of 19.5\% compared to 15.9\% with the Loquacious Set. \\

\noindent\textbf{Comparison with Whisper.} Whisper is often considered a state-of-the-art ASR system. Unfortunately, any comparison with it is made almost irrelevant outside of pure WERs benchmarking due to the hidden training data. We also cannot evaluate the models ourselves as it remains unclear if the training material contained licensed or copyrighted data. Hence, we can only compare to literature numbers on English fine-tuned models. Let us focus on the results reported for Whisper models. On the test-other of Librispeech, Whisper Small.en (244M parameters) obtains a WER of 6.7\% while our 100M models obtains 5.7\%. It takes Whisper 3 times more parameters (770M) to reach the same 5.7\% while our largest model, at 480M parameters, obtains 4.6\%. The latter WER is not even matched by the largest Whisper v2 model with 1.5B parameters at 5.1\%. Finally, this same 1.5B model obtains 7.5\% of WER on VoxPopuli while we reach 6.9\% with 480M parameters and 7.6\% with 250M, i.e. with 6 times fewer parameters.

\section{Conclusion}
The Loquacious Set offers 25,000 hours of normalised transcribed and diverse English speech recognition data available for academics, researchers in the industry and commercial use. The dataset is easy to reproduce thanks to the SpeechBrain released source code and can be loaded in a single line of code. It introduces a new common ground for benchmarking novel speech recognition systems, but on real-world conditions and at scale. Finally, the provided SpeechBrain conformer recipes exhibit state-of-the-art word error rates, while requiring as few as four GPUs, and even beat systems whose training data is larger and mostly undocumented. 

\bibliographystyle{IEEEtran}
\bibliography{bibliography}

\end{document}